\documentclass[11pt]{article}

\usepackage[a4paper,margin=1in]{geometry}
\usepackage[T1]{fontenc}
\usepackage[utf8]{inputenc}
\usepackage{lmodern}
\usepackage{microtype}
\usepackage{amsmath,amssymb}
\usepackage{graphicx}
\usepackage{subcaption}
\usepackage{placeins}
\usepackage{booktabs}
\usepackage{array}
\usepackage{tabularx}
\usepackage{siunitx}
\usepackage{xcolor}
\usepackage[numbers,sort&compress]{natbib}
\usepackage{xurl}
\usepackage[colorlinks=true,linkcolor=blue,citecolor=blue,urlcolor=blue]{hyperref}

\newcommand{\fastfluent}{FastFluent}
\newcommand{\fluent}{Fluent}

\newcommand{\keywords}[1]{\par\vspace{0.75em}\noindent\textbf{Keywords:} #1\par}
\newcolumntype{Y}{>{\raggedright\arraybackslash}X}
\newcolumntype{P}[1]{>{\raggedright\arraybackslash}p{#1}}

\usepackage{pdfpages}

\title{FastAgent: A Physics-Aware FastFluent-to-Fluent Workflow for Rapid CFD Screening and Industrial Steam Dewaxing Design}

\author{
  Chen YanKai$^{2,\dagger}$,
  Yu QiuHeng$^{1,3,\dagger}$,
  Yuan Man$^{2}$,
  Zou Ting$^{2}$,\\[0.2em]
  Wu Lin$^{1,3}$,
  He Jian$^{1,3}$,
  Zhang QiongYuan$^{1,3,*}$,
  Jing Tao$^{2,*}$,
  and Xu QingYan$^{2,*}$\\[0.5em]
  $^{1}$State Key Laboratory of Clean and Efficient\\
  Turbomachinery Power Equipment, Deyang 618000, China\\
  $^{2}$School of Materials Science and Engineering,\\
  Tsinghua University, Beijing, China\\
  $^{3}$Dongfang Electric Corporation Dongfang Turbine Co., Ltd.,\\
  Deyang 618000, China\\[0.4em]
  $^{\dagger}$These authors contributed equally to this work.\\[0.4em]
  Contact: Chen YanKai, \href{mailto:chenyankai02@gmail.com}{\nolinkurl{chenyankai02@gmail.com}}
}

\date{}

\begin{document}

\maketitle

\begin{abstract}
Industrial computational fluid dynamics (CFD) becomes costly when complex geometries, long transients and coupled physics must be evaluated across large design spaces. We introduce FastAgent, a physics-aware campaign controller built around \fastfluent{}, a lightweight numerical engine for rapid physical screening. \fastfluent{} supplies stage-specific fields and diagnostics; FastAgent allocates physical models, numerical resolution and simulation horizons by the active quantities of interest (QoIs), while returned \fluent{} discrepancies update subsequent decisions. Across six public CFD benchmark campaigns, the framework reduces normalised worst-QoI error by 5.68--99.14\% and accelerates decision-relevant computation by factors of 1.53--15.36 relative to a matched Generic Agent control.

Applied to high-pressure steam dewaxing in investment casting, FastAgent identifies nonuniform condensation heating as the dominant startup mechanism and separates rapid risk-window localisation from high-fidelity melt-completion confirmation. The rapid route requires 81.39\% fewer cell-time updates than a fine-grid, small-time-step reference. Structural analysis localises fracture at the observed blade edge. Screening 131,072 outlet layouts and selectively evaluating candidates with \fluent{} and Abaqus identifies Sparse-6, which reduces simulated crack opening by 96.0\% and connected crack extent by 66.7\% relative to the Original design. These results demonstrate how rapid numerical evidence and adaptive fidelity control connect large CFD search spaces to mechanism-guided industrial design.
\end{abstract}


\section{Introduction}
Industrial computational fluid dynamics (CFD) is particularly demanding when complex geometries and transient multiphysics must be evaluated repeatedly across design alternatives. In aircraft aerodynamic design, geometry changes require repeated preparation, meshing and boundary assignment \citep{beall2004comparison,slotnick2014cfd,karman2017mesh}. Off-design turbofan transients require time-resolved calculations of evolving flow and thermal states \citep{slotnick2014cfd}. In investment casting, high-pressure steam dewaxing couples condensation, wax melting, drainage and ceramic-shell loading over distinct time scales \citep{jones2003moisture,jones2004vessel,jones2004wax,lee2016understanding}. Such studies form CFD campaigns, comprising coordinated calculations across candidates and process stages in which the target quantities of interest (QoIs), required fidelity and computational cost must be considered together.

High-fidelity solvers provide detailed predictions for selected configurations, while low-cost models broaden exploration. Classical multi-fidelity methods combine these evaluations to manage accuracy and cost \citep{peherstorfer2018multifidelity}. In multi-stage applications, accuracy is specific to a physical question. A calculation may resolve the timing of a critical event yet miss the process endpoint, and global performance may conceal a local failure mechanism. Efficient exploration therefore requires fidelity allocation that changes with the active QoI and uses high-fidelity discrepancies to revise subsequent calculations.

Language-model agents coordinate scientific workflows through reasoning and tool use \citep{yao2023react,schick2023toolformer,wu2024autogen,yang2024sweagent,boiko2023coscientist,bran2024chemcrow,holland2024processagent,zhao2026shopflooragent,jadhav2025llm3dprint}. CFD-specific systems automate case construction, meshing, solver execution, error correction and post-processing, with extensions to sensitivity analysis and optimisation \citep{pandey2025openfoamgpt,chen2025metaopenfoam2,feng2026openfoamgpt2,yue2026foamagent,fan2026chatcfd,chen2026optmetaopenfoam}. Structured knowledge, physical-model selection and field-level assessment further improve execution and physical fidelity \citep{yue2026foamagent,fan2026chatcfd,somasekharan2026cfdllmbench}. Recent studies also extend CFD agents to physics-verified hypothesis search \citep{somasekharan2026aicfdscientist} and surrogate-assisted multi-fidelity engineering workflows \citep{son2026aei}. However, current systems primarily automate prescribed solver workflows or individual analysis steps. They rarely treat the physical model, numerical resolution, simulation horizon and candidate escalation as coupled decisions that must change with the active QoI. Repeated reliance on high-fidelity solvers also limits their ability to explore combinatorial design spaces and long transients spanning disparate time scales. FastAgent addresses this campaign-level gap by coupling a purpose-built rapid physics engine with QoI-aware control and discrepancy feedback. This combination generates comparable numerical evidence across many candidates, reserves high-fidelity computation for decision-critical states and uses confirmed discrepancies to revise subsequent decisions.

We introduce FastAgent, a physics-aware controller built around \fastfluent{}, a lightweight numerical engine developed for rapid CFD screening. \fastfluent{} brings lattice-Boltzmann flow calculations, conservative finite-volume balances and reduced-order thermal phase-change screening into a common field--QoI--diagnostic representation. This representation supplies physical responses, conservation information and computational cost for candidate comparison. FastAgent treats the physical model, numerical resolution and simulation horizon as QoI-dependent decisions and selects \fluent{} calculations for high-fidelity confirmation. Returned discrepancies update candidate ranking and subsequent fidelity allocation. The framework combines three contributions, comprising a purpose-built rapid physics engine, QoI-specific control across process stages and discrepancy-updated design selection. \fastfluent{} supplies the computational reach for broad exploration, and FastAgent turns that evidence into an adaptive calculation path. Six public CFD benchmark campaigns quantify the resulting accuracy, efficiency and decision quality.

Steam dewaxing of hollow turbine-blade investment shells exposes why these capabilities must work together. Rapid heating promotes wax removal but also drives constrained expansion against the ceramic shell, while melting and outlet-connected drainage redistribute and release the load \citep{jones2003moisture,jones2004vessel,jones2004wax,lee2016understanding,piwonka2000modeling,sabau2003material,mukhtarkhanov2025novel}. The computational difficulty is stage-dependent. The opening 0--4.1 s requires discrimination between steam momentum and condensation-driven heating, the 0--420 s cycle requires different accuracy for risk-window timing and melt completion, and outlet design couples global drainage to local blade-edge fracture. In the present design space, 17 binary outlet positions generate 131,072 layouts, making uniform high-fidelity evaluation an expensive basis for exploration. \fastfluent{} screens the full space, while FastAgent directs selective \fluent{} and Abaqus evaluation and incorporates the identified fracture mechanism into layout selection. This combination links physical-model selection, QoI-dependent simulation and fracture-aware optimisation within one campaign. The selected Sparse-6 design reduces simulated crack opening by 96.0\% relative to the Original design, demonstrating how campaign control translates large-scale computation into a concrete design intervention.

\section{Results}
\subsection{FastAgent allocates CFD fidelity through physics-aware campaign control}
FastAgent couples a campaign decision policy to \fastfluent{}'s rapid numerical evidence (Fig.~\ref{fig:agent-evidence-loop}a). The campaign specifies geometry, physical assumptions, candidate parameters, target QoIs and a computational budget. As results accumulate, FastAgent updates four coupled choices, comprising the physical model, numerical resolution, simulation horizon and candidates selected for high-fidelity evaluation.

\fastfluent{} is the computational engine for rapid exploration. Its stage-specific formulations return physical fields, QoIs, conservation diagnostics and computational cost in a common evidence representation. FastAgent uses these outputs to revise or rescreen candidates, extend simulations and select \fluent{} calculations. \fluent{} comparisons quantify the discrepancies that update later ranking and fidelity assignments, while Abaqus resolves the structural response for fracture-related decisions. \fastfluent{} provides the low-cost calculations needed to explore many alternatives, and FastAgent determines how that evidence changes the next calculation.

\begin{center}
  \begin{minipage}{0.98\textwidth}
    \centering
    \includegraphics[width=\textwidth]{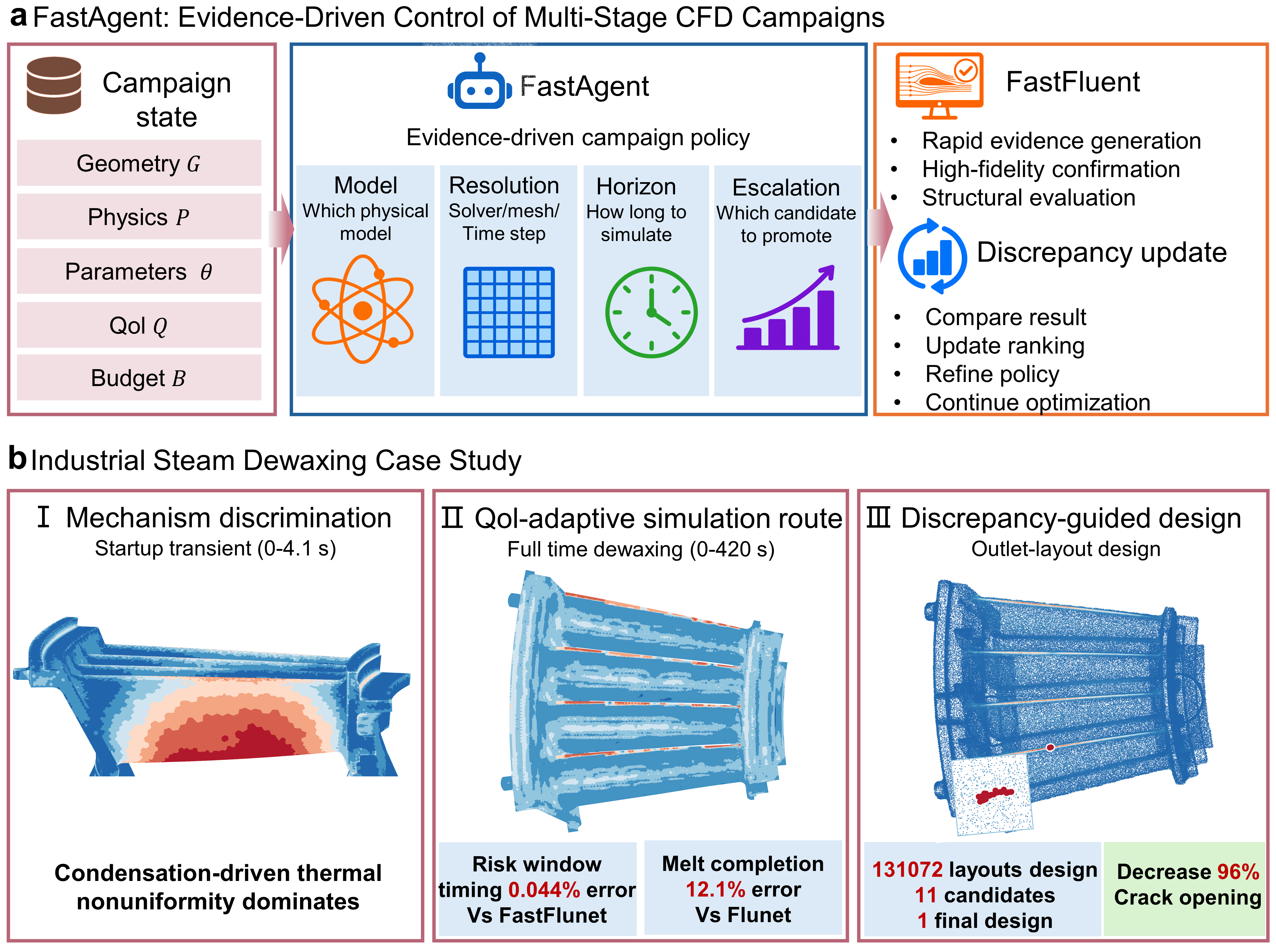}
    \captionsetup{hypcap=false}
    \captionof{figure}{\textbf{FastAgent couples evidence-driven CFD campaign control with stage-specific industrial decisions.} \textbf{a}, A campaign state is converted into decisions over the physical model, numerical resolution, simulation horizon and escalation of decision-critical candidates. \fastfluent{} generates rapid physical evidence, selected cases are confirmed with \fluent{} or evaluated structurally with Abaqus, and returned discrepancies update subsequent ranking and campaign decisions. \textbf{b}, The steam-dewaxing application instantiates the same policy through startup mechanism discrimination, QoI-adaptive simulation routing during the full cycle, and discrepancy-guided outlet-layout screening from 131,072 \fastfluent{} calculations to 11 \fluent{} confirmations, three Abaqus evaluations and the Sparse-6 design.}
    \label{fig:agent-evidence-loop}
  \end{minipage}
\end{center}
\FloatBarrier

\noindent\begin{minipage}{\textwidth}
\setlength{\parindent}{1.5em}
In steam dewaxing, this coupling links three decisions (Fig.~\ref{fig:agent-evidence-loop}b). Startup analysis selects the condensation treatment needed to distinguish thermal and momentum loading. Full-cycle analysis assigns rapid calculation to risk-window timing and \fluent{} confirmation to melt completion. Outlet-layout design combines complete \fastfluent{} screening with \fluent{}-based ranking updates and Abaqus fracture evaluation. The same numerical--decision coupling thus connects mechanism identification to design selection. The campaign policy and \fastfluent{} formulations are defined in Methods Sections~3.1--3.2.
\end{minipage}

\subsection{FastAgent improves decision quality and efficiency across six public CFD benchmark campaigns}
Six public CFD benchmark campaigns evaluate complementary aspects of FastAgent. The laminar channel tests numerical correction, the backward-facing step tests transient diagnosis and rescreening, the heated channel tests thermal-QoI transfer, and the lid-driven cavity tests full-field confirmation. The Ghia cavity and Sch\"afer--Turek cylinder cases assess discrepancy-driven correction and adaptive stopping. Together, the suite, comprising four capability benchmarks and two literature reproductions, quantifies physical accuracy, computation time and agent-level decision quality (Table~\ref{tab:six-case-headline-results} and Supplementary Tables~S1--S3 and S10--S11).

\begin{figure}[htbp]
  \centering
  \includegraphics[width=\textwidth]{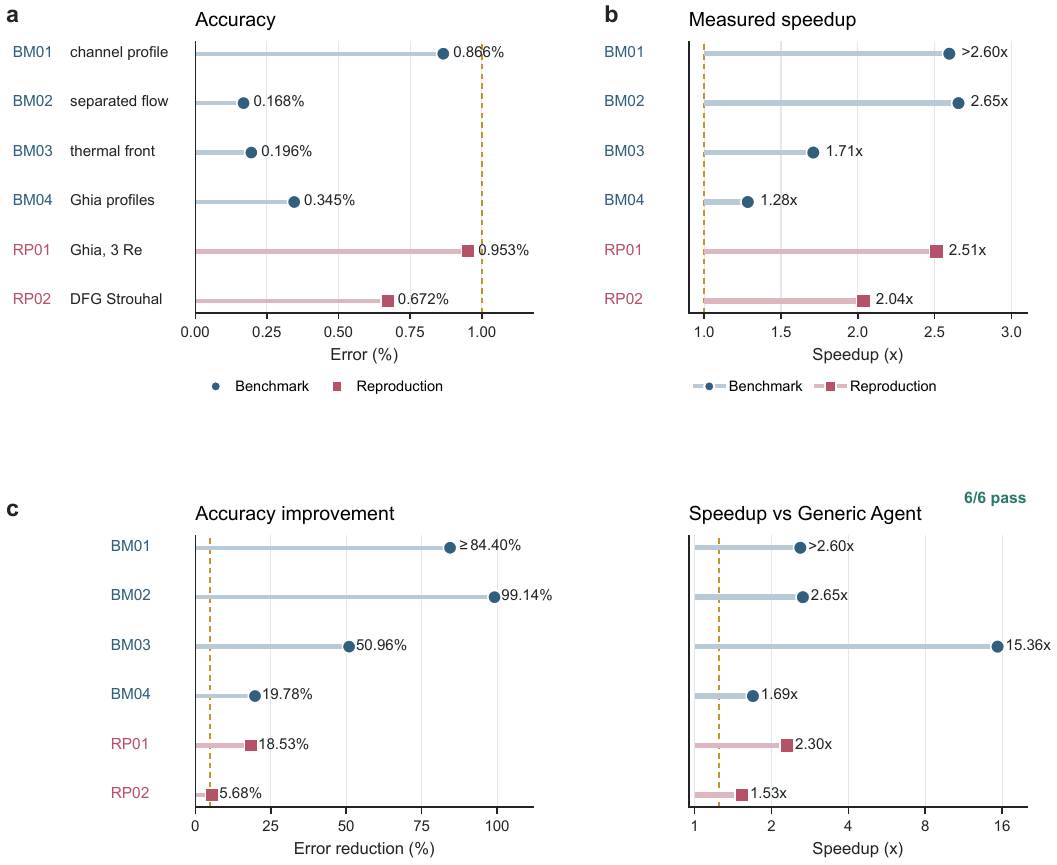}
  \caption{\textbf{FastAgent validation across six public CFD benchmark campaigns.} \textbf{a}, Representative error metrics or diagnostic deviations for the six cases. \textbf{b}, Measured speedups relative to expert-defined solver baselines. \textbf{c}, Reduction in normalised worst-QoI error and decision-relevant computational speedup relative to the matched Generic Agent control. Dashed lines mark the 5\% accuracy-improvement and $1.25\times$ speedup criteria. All six cases satisfy both. Labels BM01--BM04 denote the laminar channel, backward-facing step, heated channel and lid-driven cavity. RP01 and RP02 denote the Ghia cavity and Sch\"afer--Turek cylinder reproductions. The laminar-channel values are lower bounds, and the cylinder comparisons begin from a shared 4.0 s checkpoint.}
  \label{fig:accuracy-efficiency}
\end{figure}

Figure~\ref{fig:accuracy-efficiency}b compares computational cost with fixed \fluent{} schedules, while Fig.~\ref{fig:accuracy-efficiency}c compares the FastAgent--\fastfluent{} workflow with the Generic Agent workflow. Matched-evidence replay measures stagewise decisions using the same progressively revealed CFD results (Supplementary Tables~S10 and S11).

In the replay, FastAgent satisfied all stagewise criteria in six campaigns, compared with five for the Generic Agent (19/19 versus 18/19 prescribed actions). At the backward-facing-step stationarity check, continued evolution of the reattachment length prompted FastAgent to rescreen; the Generic Agent continued from the same evidence.

\begin{table}[htbp]
  \centering
  \caption{\textbf{FastAgent decisions and principal evidence across six CFD campaigns.} The cases test numerical correction, transient diagnosis, QoI transfer, full-field confirmation, discrepancy-driven refinement and adaptive stopping. Reported speedups use the case-specific solver baselines in Fig.~\ref{fig:accuracy-efficiency}b.}
  \label{tab:six-case-headline-results}
  {\footnotesize
  \setlength{\tabcolsep}{4pt}
  \renewcommand{\arraystretch}{1.22}
  \begin{tabularx}{\textwidth}{@{}P{2.25cm}P{2.45cm}YYY@{}}
    \toprule
    Case & Physical test & Agent/\fastfluent{} decision & Key result & Supported conclusion \\
    \midrule
    Laminar channel & Refinement and mass balance & Diagnose and repair the inlet before high-fidelity confirmation & RMS $0.008656$; speedup $>2.595\times$ & Numerical correction \\
    Backward-facing step & Separated-flow reattachment & Screen three tiers and return a non-stationary result & $2.655\times$ acceleration; snapshot deviation $0.168\%$ & Rapid screening with diagnostic return \\
    Heated channel & Blocked thermal transport & Correct porous mapping through front and dose monitors & Errors $0.1956\%/0.5847\%$; speedup $1.710\times$ & Stage-specific QoI transfer \\
    Lid-driven cavity & Full-field confirmation & Transfer the velocity field and correct numerical settings & Ghia RMS $0.003452$; speedup $1.283\times$ & High-fidelity full-field confirmation \\
    Ghia cavity reproduction & $Re=100$--$1000$ profiles & Screen nine candidates and confirm three & Mean RMS $0.0095285$; speedup $2.512\times$ & Discrepancy-driven correction \\
    Sch\"afer--Turek cylinder & Unsteady wake at $Re=100$ & Stop after consecutive multi-QoI criteria are satisfied & Strouhal error $0.672\%$; speedup $2.037\times$ & Adaptive confirmation \\
    \bottomrule
  \end{tabularx}}
\end{table}

\FloatBarrier

The laminar channel tests numerical correction. Grid refinement reduces the \fastfluent{} velocity error from $2.78\times10^{-2}$ to $4.34\times10^{-4}$ while preserving mass balance. After inlet correction, high-fidelity calculation with \fluent{} gives a normalised outlet-profile RMS error of 0.0087, and the complete calculation achieves a speedup greater than 2.60 relative to the corresponding \fluent{} control. The backward-facing step then tests whether scalar agreement is supported by the evolving flow. The selected snapshot agrees with the reference reattachment position to 0.168\%. Continued monitoring identifies the remaining temporal variation, and FastAgent returns the case for rescreening. The adaptive calculation achieves a 2.65-fold speedup over the three-tier \fluent{} sweep (Supplementary Fig.~S1 and Tables~S1 and S3).

The heated channel connects the validation suite to the thermal physics used in steam dewaxing. The clear-channel outlet temperature agrees with \fluent{} to 0.005\%, while the obstructed case shows that thermal-front position and blocked-zone dose provide more discriminating local measures. Corrected porous mapping limits their errors to 0.20\% and 0.58\%, with a 1.71-fold speedup. The lid-driven cavity then tests full-field high-fidelity confirmation. The confirmed velocity field reaches a Ghia-profile RMS error of 0.0035 and preserves the vortex centre. The complete calculation reduces wall-clock time by a factor of 1.28 (Supplementary Fig.~S2 and Tables~S1 and S3).

The Ghia cavity reproduction extends the test over three conditions \citep{ghia1982high}, with $Re=100$, 400 and 1000. Discrepancy-driven anchor selection reduces the aggregate profile RMS error from 0.121 to 0.0095 and limits high-fidelity confirmation to three of nine candidates, giving a 2.51-fold speedup. This case demonstrates that rapid screening can identify where published-profile agreement requires targeted correction (Supplementary Fig.~S3 and Tables~S2 and S3).

The Sch\"afer--Turek cylinder reproduction at $Re=100$ tests adaptive control of an unsteady simulation \citep{schaefer1996benchmark}. The frequency-calibrated configuration predicts $St=0.300$, a 0.67\% error, and satisfies the prescribed frequency and force-statistics criteria. Consecutive successful evaluations terminate the high-fidelity continuation with a 2.04-fold speedup over the fixed-budget control. An amplitude-calibrated variant improves the force ranges while retaining the shedding frequency, exposing a multi-QoI trade-off that favours the frequency-calibrated result (Supplementary Fig.~S4 and Tables~S2 and S3).

The representative errors and diagnostic deviations shown in Fig.~\ref{fig:accuracy-efficiency}a remain below 1\%, with measured speedups of 1.28--2.65 against the fixed \fluent{} schedules. Relative to the Generic Agent workflow, the FastAgent--\fastfluent{} workflow reduces normalised worst-QoI error by 5.68--99.14\% and accelerates decision-relevant computation by factors of 1.53--15.36. These results establish the numerical correction, rescreening and adaptive confirmation capabilities combined in the following steam-dewaxing study.
\FloatBarrier

\subsection{FastAgent-guided startup screening identifies thermal nonuniformity as the dominant early mechanism}

FastAgent formulated startup dewaxing as a mechanism-discrimination problem. It asked whether direct steam momentum or spatially nonuniform condensation heating governed the shell response before substantial wax softening, expansion, melting and drainage. FastAgent isolated the first $0$--$4.1$ s of the measured pressure history and evaluated condensation treatment, impingement orientation, distance and inlet configuration using stage-specific thermal and crack-driving QoIs \citep{jones2004vessel,jones2004wax,minkowycz1966condensation} (Supplementary Fig.~S5 and Table~S4).

FastAgent first compared three condensation treatments for the startup simulation. A non-condensing calculation raised the chamber-mean gas temperature to approximately $255\,^{\circ}$C at 4 s, whereas both condensation-aware treatments limited it to $170$--$180\,^{\circ}$C, consistent with steam saturation and published autoclave thermal profiles \citep{jones2004vessel,jones2004wax,wagner2000iapws}. The equivalent-condensation treatment also reproduced the imposed pressure rise within 3.9\%, so FastAgent retained it for the orientation and distance campaign. The resulting \fluent{} fields resolve jet establishment, redistribution and progressive heating through 4.1 s (Supplementary Figs.~S6 and S7). Similar bulk histories produce distinct jet paths and surface footprints, establishing the need for surface-resolved measures of local shell exposure.

\begin{figure}[p]
  \centering
  \makebox[\textwidth][c]{\includegraphics[width=1.08\textwidth]{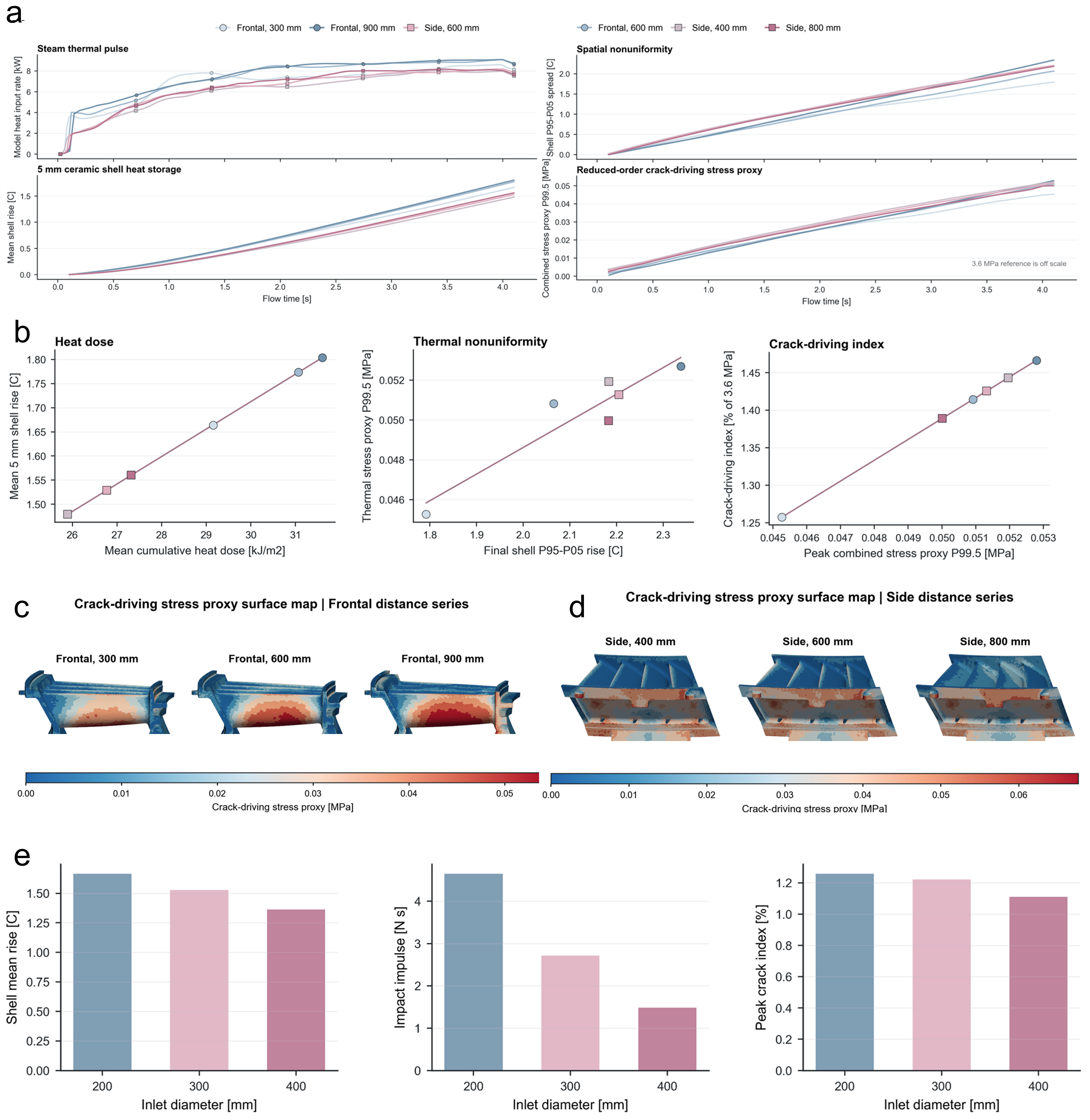}}
  \caption{\textbf{FastAgent-guided startup screening identifies thermal nonuniformity as the dominant early mechanism.} \textbf{a}, Time-resolved steam heat input, equivalent 5 mm shell heating, spatial temperature spread and combined stress proxy for six cases varying orientation and distance. \textbf{b}, Cross-case relations connecting heat dose to mean shell heating, temperature nonuniformity to the thermal-stress contribution and the combined stress proxy to the crack-driving index. \textbf{c,d}, Final frontal and side stress-proxy maps. \textbf{e}, Inlet-diameter sensitivity of shell heating, impact impulse and peak crack-driving index.}
  \label{fig:early-steam-shock-response}
\end{figure}

The post-processing model converted transient wall heat flux into cumulative heat dose and an equivalent response for a 5 mm ceramic shell. Total heat input approaches 8--9 kW near the end of the interval (Fig.~\ref{fig:early-steam-shock-response}a). Mean heat dose spans 29.2--31.6 kJ m$^{-2}$ for frontal impingement and 25.9--27.3 kJ m$^{-2}$ for side impingement, producing mean shell-temperature rises of 1.48--1.80$\,^{\circ}$C. Spatial nonuniformity grows throughout the interval and reaches a maximum $P_{95}$--$P_{05}$ spread of 2.34$\,^{\circ}$C. Heat dose governs mean temperature rise. Spatial variation governs the thermal-stress contribution, consistent with ceramic thermal-shock theory \citep{kingery1955thermal} (Fig.~\ref{fig:early-steam-shock-response}b and Supplementary Fig.~S8).

Pressure loading exhibits a different spatial pattern. Side impingement produces the largest integrated momentum transfer, reaching an impact impulse of 27.9 N s, while the largest normal-impact contribution is only 0.0034 MPa. A common shell-response scale combines thermal nonuniformity and direct pressure loading as
\[
\sigma_{\mathrm{proxy}}=\sigma_T+\sigma_p,
\qquad
I_{\mathrm{crack}}=\frac{\sigma_{\mathrm{proxy},P_{99.5}}}{\sigma_{\mathrm{ref}}}.
\]
For the six orientation and distance cases, the maximum combined $P_{99.5}$ response is 0.053 MPa. The crack-driving indices span 1.26--1.47\% of the adopted 3.6 MPa green-shell strength scale \citep{lee2016understanding,lee2017shell} (Supplementary Table~S5). The thermal component supplies most of the combined response, and direct normal impact adds only a small increment. The maps preserve the orientation-dependent footprint on the heated surfaces (Fig.~\ref{fig:early-steam-shock-response}c,d and Supplementary Fig.~S9). A separate inlet-diameter series gives the same trend. Increasing the inlet diameter from 200 to 400 mm reduces mean shell heating, impact impulse and peak crack-driving index (Fig.~\ref{fig:early-steam-shock-response}e).

\FloatBarrier

FastAgent ranked nine startup configurations by final cavity-average temperature and local crack-driving response. Under the $170$--$180\,^{\circ}$C process target, frontal impingement through a 200 mm inlet positioned 300 mm from the pattern ranked first. This configuration reached $179.5\,^{\circ}$C and limited the peak combined response to 0.045 MPa. The dual-inlet configuration produced the largest response in the expanded set at 0.071 MPa. All nine configurations remained below 2\% of the green-shell reference strength under the adopted conversion gains. The preferred configuration also remained unchanged throughout the two-parameter gain sensitivity sweep (Supplementary Table~S6).

Thermal nonuniformity dominates the startup response, with the frontal, 300 mm, 200 mm-inlet configuration providing the preferred balance of process temperature and shell loading. This result redirects FastAgent to the subsequent wax expansion, melting and drainage stages, where shell--wax contact loading determines the dominant load path \citep{lee2016understanding,piwonka2000modeling,sabau2003material,mukhtarkhanov2025novel}.

\subsection{FastAgent-guided full-cycle fidelity allocation links dewaxing loads to blade-edge fracture}

\begin{figure}[h!]
  \centering
  \includegraphics[width=0.98\textwidth]{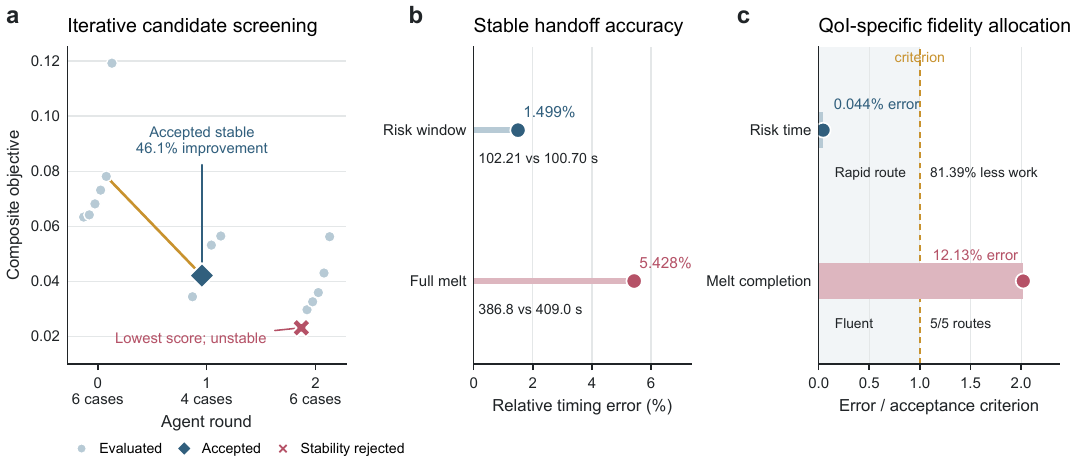}
  \caption{\textbf{FastAgent-guided full-cycle fidelity allocation separates risk localisation from melt-completion confirmation.} \textbf{a}, Three-round search of 16 candidates and selection of a stable candidate for confirmation. \textbf{b}, Timing accuracy of the selected \fastfluent{} candidate against \fluent{}. \textbf{c}, Quantity-specific evaluation of risk timing by rapid calculation and melt completion by \fluent{}.}
  \label{fig:fullcycle-agent-screening}
\end{figure}

FastAgent treated the 0--420 s process as a QoI-specific fidelity problem spanning load accumulation, melting and drainage, process completion and structural analysis. Across three screening rounds, \fastfluent{} evaluated 16 candidates through 46 rapid calculations, rejected unstable grid or time-step combinations and retained a stable candidate for high-fidelity CFD confirmation (Fig.~\ref{fig:fullcycle-agent-screening}a). Relative to the baseline candidate, the selected state reduced the weighted reduced-order objective defined in Methods from 0.0781 to 0.0421, an improvement of 46.1\%. Its risk-window and full-melt times differed from the corresponding \fluent{} results by 1.50\% and 5.43\%, respectively (Fig.~\ref{fig:fullcycle-agent-screening}b).

\FloatBarrier

For QoI-specific numerical fidelity, FastAgent next compared five spatial--temporal resolution routes around the baseline configuration. The large-time-step setting reproduced the 100.7 s \fluent{} risk-window time with 0.044\% error and the integrated energy within 5.16\%. Its 928,200 cell-time updates were 81.39\% fewer than the fine-grid, small-time-step reference, and it remained selected across all 30 neighbouring criterion combinations. Its full-melt error was 12.1\%, and all five routes in this resolution comparison exceeded the 6\% endpoint criterion. FastAgent therefore used rapid calculation to locate the risk window and reserved high-fidelity CFD for melt-completion confirmation (Fig.~\ref{fig:fullcycle-agent-screening}c and Supplementary Table~S7).

Full-cycle process timing and local loading were resolved through complementary measures. The global screening-pressure descriptor peaked at 4.47 MPa near 100.7 s, defining the timing QoI for the fidelity study (Fig.~\ref{fig:fullcycle-agent-screening}b). A spatial high-load descriptor constructed from \fluent{} temperatures and the pressure--volume--temperature (PVT) relation reached a $P_{99.9}$ peak of 3.82 MPa at 17.4 s (Supplementary Fig.~S11). Full melt occurred at 409 s, with a final average liquid fraction of 0.997.

FastAgent directed structural analysis to the early blade-edge loading interval. The PVT and shell-compliance formulation converted retained wax expansion into regional loads (Methods Section~3.6); a retained drive of 190 mm$^3$ produced a maximum regional load of 0.0275 MPa at 8.38 s (Supplementary Table~S8). A 0.1 s rapid scan bracketed the initial connected edge clusters, and nonlinear XFEM resolved the corresponding events at approximately 8.38, 8.44 and 8.53 s (Supplementary Fig.~S10). The 8.3--9.1 s structural interval captures early local fracture evolution, while the 100.7 s descriptor locates the later global screening-risk window.

\begin{figure}[p]
  \centering
  \includegraphics[height=0.865\textheight,keepaspectratio]{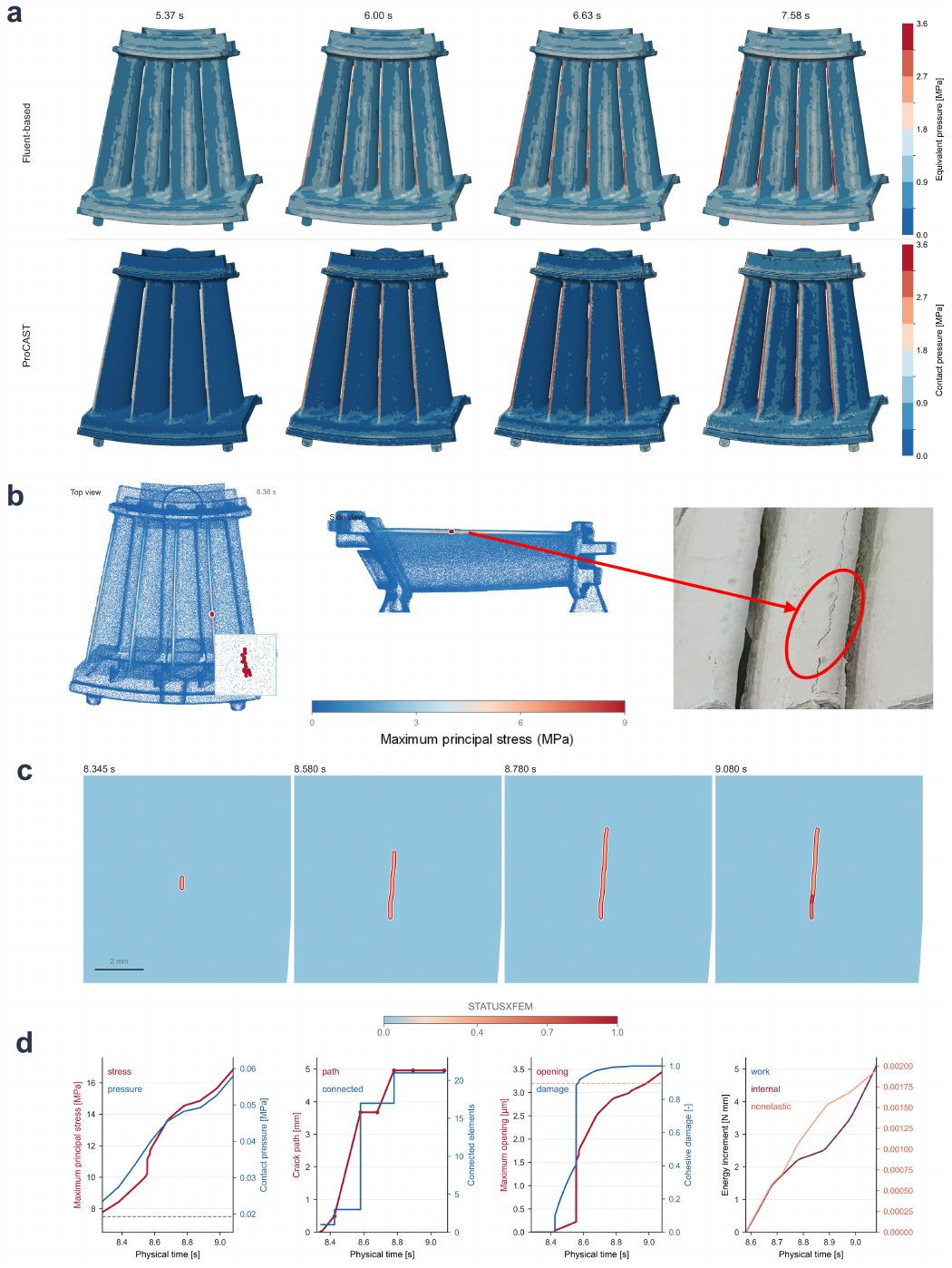}
  \caption{\textbf{FastAgent-guided structural analysis links dewaxing loads to blade-edge fracture.} \textbf{a}, Thermal and pressure states converted into regional shell loading and compared with independently calculated ProCAST contact pressure. \textbf{b}, Resulting maximum-principal-stress localisation and the observed blade-edge shell crack. \textbf{c}, XFEM crack initiation and edge-aligned growth. \textbf{d}, Mechanical histories describing stress, crack connectivity, opening, damage and energy.}
  \label{fig:dewaxing-pressure-fracture}
\end{figure}

The reconstructed regional load intensified along narrow blade-edge bands between 5.37 and 7.58 s (Fig.~\ref{fig:dewaxing-pressure-fracture}a). Independently calculated ProCAST contact pressures showed the same blade-edge localisation and provided a cross-solver spatial check. By 8.38 s, the mapped load placed the maximum-principal-stress concentration at a blade edge (Fig.~\ref{fig:dewaxing-pressure-fracture}b). Crack initiation occurred at approximately 8.35 s, followed by cohesive opening at 8.43 s. The connected path approached 5.0 mm and stabilised after 8.78 s, while the maximum opening reached 3.44$\,\mu\mathrm{m}$ at 9.08 s (Fig.~\ref{fig:dewaxing-pressure-fracture}c,d). Local separation continued after the path length stabilised. By 8.58 s, 39 damage elements lay on blade edges and the root region remained undamaged. The calculation localises fracture consistently with the observed blade-edge crack region.

The \fluent{}-derived regional load, independent ProCAST contact pressures and Abaqus fracture response localise the same blade-edge failure mode. FastAgent carries this mode forward as the local structural objective for outlet-layout selection.

\FloatBarrier

\subsection{FastAgent-guided outlet-layout design balances drainage performance and fracture risk}

\begin{figure}[p]
  \centering
  \makebox[\textwidth][c]{\includegraphics[width=1.12\textwidth]{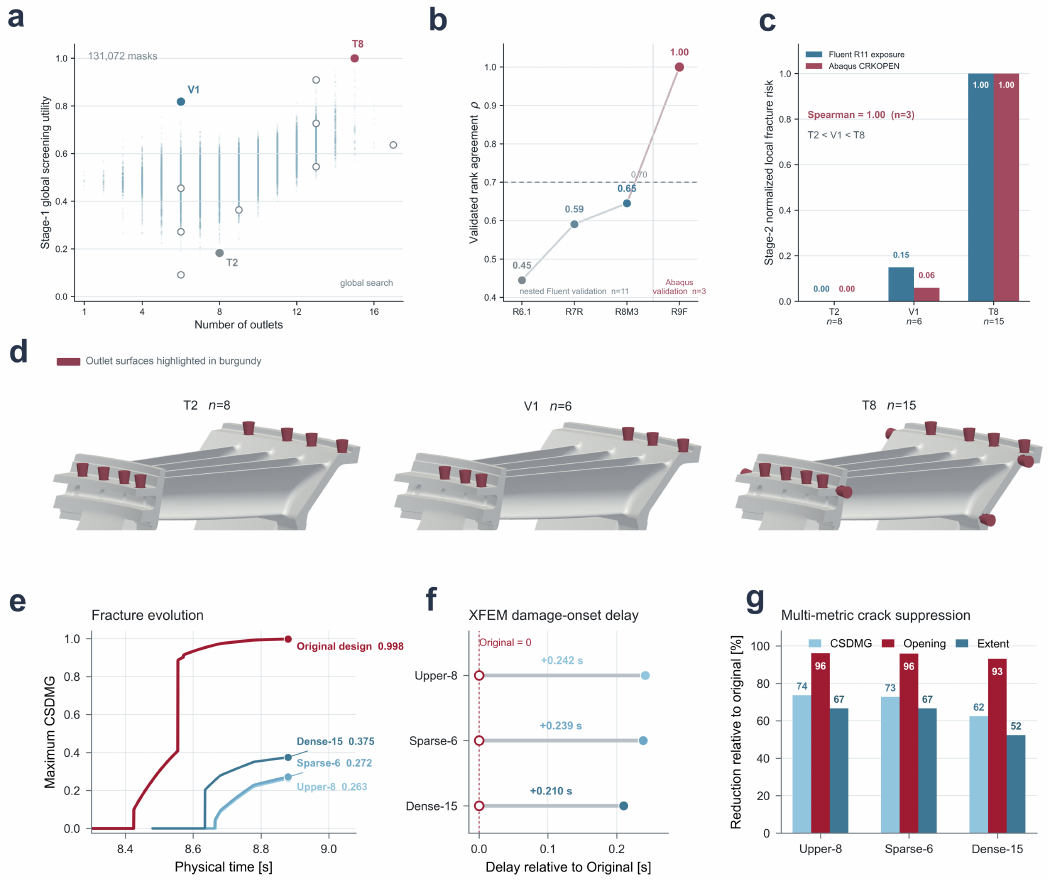}}
  \caption{\textbf{FastAgent-guided outlet-layout design balances drainage performance and local fracture safety.} \textbf{a}, Complete \fastfluent{} design-space screening followed by selective high-fidelity CFD and structural analysis. \textbf{b}, Global ranking agreement across the eleven \fluent{}-evaluated layouts for the baseline, rank-calibrated and physics-informed stages (R6.1, R7R and R8M3). The fracture-aware point (R9F) instead reports local-risk agreement across the three structurally evaluated layouts. \textbf{c}, Consistent local-risk ordering from \fluent{} critical-region exposure and Abaqus crack opening; R11 denotes the exposure metric. \textbf{d}, Upper-8, Sparse-6 and Dense-15 layouts, labelled T2, V1 and T8 in the figure. \textbf{e--g}, Cohesive damage, activation delay and reductions in damage, opening and connected extent relative to the Original design.}
  \label{fig:agent-outlet-design}
\end{figure}

FastAgent converted the blade-edge failure mechanism into a two-objective outlet-layout design problem. Global drainage utility rewarded drainage completion and reduced shell-load exposure, while local fracture safety retained the critical blade-edge response identified in Section~2.4. Seventeen candidate outlet positions generated $2^{17}=131{,}072$ layouts, all screened by \fastfluent{} (Fig.~\ref{fig:agent-outlet-design}a). FastAgent selected eleven representative layouts for high-fidelity CFD evaluation with \fluent{} and retained Upper-8, Sparse-6 and Dense-15 for structural comparison with Abaqus. \fastfluent{} establishes the global design map; \fluent{} and Abaqus resolve drainage and fracture behaviour for the selected layouts.

Each \fluent{} comparison returned a discrepancy to FastAgent and updated the screening model. Rank agreement across the eleven comparable layouts increased from 0.445 for the baseline model to 0.591 after rank calibration and 0.645 after physics-informed refinement (Fig.~\ref{fig:agent-outlet-design}b). The blade-edge failure identified in Section~2.4 then introduced critical-region pressure exposure as a local selection objective. Across the three layouts examined structurally, \fluent{} exposure and Abaqus crack opening produced the same risk order, with Spearman's rank correlation equal to 1.00 (Fig.~\ref{fig:agent-outlet-design}c). FastAgent retained global drainage utility and local fracture safety as separate objectives so that broad exploration and final structural selection remained physically aligned.

The three selected layouts occupied distinct positions in this global and local design space (Fig.~\ref{fig:agent-outlet-design}d). Their critical-region pressure exposures were 0.0133, 0.0134 and 0.0138 MPa s for Upper-8, Sparse-6 and Dense-15, respectively. Maximum crack openings followed the same order at 0.115, 0.121 and 0.204$\,\mu\mathrm{m}$. Upper-8 minimised local fracture response and provided the weakest global drainage utility among the selected designs. Dense-15 maximised drainage utility and increased critical-edge exposure and opening. Sparse-6 preserved strong drainage performance while remaining mechanically close to Upper-8. After normalisation of global utility and local fracture safety, a weight sweep retained Sparse-6 as the highest-ranked design for global-utility weights from 0.075 to 0.765 in increments of 0.005. FastAgent therefore selected Sparse-6 as a stable balance between drainage and local shell protection.

Matched XFEM calculations quantified the structural benefit relative to the Original design. At the common endpoint of 8.88 s, the Original design reached an opening of 3.00$\,\mu\mathrm{m}$ and a 21-element connected extent (Fig.~\ref{fig:agent-outlet-design}e). Sparse-6 reduced cohesive damage, opening and connected extent by 72.8\%, 96.0\% and 66.7\%, respectively. Upper-8 produced similar reductions, while Dense-15 reduced the same quantities by 62.5\%, 93.2\% and 52.4\% (Fig.~\ref{fig:agent-outlet-design}g). Damage activation was delayed by approximately 0.24 s for Upper-8 and Sparse-6 and by 0.21 s for Dense-15 (Fig.~\ref{fig:agent-outlet-design}f). All three layouts retained blade-edge localisation, avoided root cracking and formed shorter connected paths than the Original design (Supplementary Fig.~S13 and Supplementary Table~S9).

Critical-region exposure and crack opening exhibit a nonlinear relationship. Dense-15 carries 3.8\% more exposure than Upper-8 but has a 76.4\% larger maximum opening. This amplification makes local fracture response decisive alongside global drainage utility. \fastfluent{} provides full-space exploration, and FastAgent integrates \fluent{} discrepancies and Abaqus fracture evidence to select Sparse-6. The resulting design combines broad numerical screening with a mechanism-based reduction in blade-edge damage.

\FloatBarrier

\section{Methods}

\subsection{FastAgent campaign representation and decision policy}

Each campaign is represented by
\begin{equation}
C = \{G, P, \Theta, Q, B\},
\label{eq:campaign_specification}
\end{equation}
where $G$ defines the geometry and fluid domain, $P$ contains the physical assumptions, $\Theta$ spans the candidate parameter space, $Q$ specifies the target quantities of interest (QoIs) and $B$ sets the computational budget. FastAgent decomposes this specification into physically distinct stages. At iteration $r$, the active candidate set $\Theta_r$, remaining budget $B_r$ and accumulated evidence history $H_r$ define the next decision context.

For candidate $k$, \fastfluent{} converts the calculated fields into stage-specific decision variables through observable operators $\mathcal{M}_j$:
\begin{equation}
\begin{aligned}
\Phi_{k,r} &= \{\mathbf{u}, p, T, \alpha, f_l\}_{k,r}, \\
q_{k,r,j} &= \mathcal{M}_j[\Phi_{k,r}], \\
E_{k,r} &= \{\Phi_{k,r}, q_{k,r}, \Pi_{k,r}, \eta_{k,r}, c_{k,r}\}, \\
d_{k,r} &= D_{\mathrm{FA}}(E_{k,r}, Q, B_r, H_r).
\end{aligned}
\label{eq:evidence_object}
\end{equation}
The field state $\Phi_{k,r}$ contains velocity, pressure, temperature, phase indicator and liquid fraction as required by the case. The evidence object $E_{k,r}$ adds the QoI vector $q_{k,r}$, physical-admissibility and numerical-quality criteria $\Pi_{k,r}$, conservation diagnostics $\eta_{k,r}$ and measured computational cost $c_{k,r}$. The decision function $D_{\mathrm{FA}}$ returns revision, rescreening, continuation, high-fidelity confirmation, stopping or rejection.

FastAgent checks specification completeness, physical admissibility, numerical reliability and QoI coverage in that order. Incomplete or inadmissible cases trigger revision; unresolved sensitivity, discrepancy or temporal evolution triggers rescreening or continuation. Computational cost enters candidate ranking after these requirements have been met.

Let $K_r$ denote the candidates screened in iteration $r$ and $P_r$ the subset selected for high-fidelity confirmation. Selection and discrepancy evaluation follow
\begin{equation}
\begin{aligned}
P_r &= \operatorname{TopK}_{k \in K_r} S(E_{k,r}, Q, H_r),
\qquad |P_r| \ll |K_r|, \\
\sum_{k \in K_r} c^{\mathrm{FF}}_{k,r}
+ \sum_{k \in P_r} c^{\mathrm{HF}}_{k,r}
&\leq B_r, \\
\epsilon_{k,r,j}
&= \frac{\left|q^{\mathrm{HF}}_{k,r,j}-q^{\mathrm{FF}}_{k,r,j}\right|}{\tau_j}.
\end{aligned}
\label{eq:budgeted_confirmation}
\end{equation}
The ranking operator $S$ first applies the physical and numerical gates, then orders admissible candidates by active-QoI coverage, normalised discrepancy and cost. The tolerance $\tau_j$ places heterogeneous QoIs on a common scale. Confirmation evidence may come from high-fidelity CFD with \fluent{} or from analytical and published reference data. Returned discrepancies update $H_r$, candidate ranking and later fidelity assignment. The campaign terminates when all active validity and QoI criteria are satisfied, including any prescribed persistence requirement for transient cases. Supplementary Tables~S10 and S11 provide the detailed decision criteria and stagewise records.

\subsection{FastFluent governing equations and numerical formulations}

\fastfluent{} generates campaign evidence with three stage-specific numerical formulations. These comprise lattice-Boltzmann flow calculations on regular domains, conservative finite-volume balances on unstructured domains and reduced-order thermal phase-change screening for full-cycle dewaxing. Each formulation supplies fields, QoIs, conservation diagnostics and computational cost to the common evidence representation in Eq.~\ref{eq:evidence_object}. The dewaxing screening model is specified in Section~3.5. The governing physical components are
\begin{equation}
\begin{aligned}
\frac{\partial \rho}{\partial t}+\nabla\!\cdot(\rho\mathbf{u})&=0,\\
\frac{\partial (\rho\mathbf{u})}{\partial t}
+\nabla\!\cdot(\rho\mathbf{u}\otimes\mathbf{u})
&=-\nabla p+\nabla\!\cdot\boldsymbol{\tau}
+\rho\mathbf{g}+\mathbf{f}_{\sigma}+\mathbf{S}_{m},\\
\frac{\partial (\rho H)}{\partial t}
+\nabla\!\cdot(\rho\mathbf{u}H)
&=\nabla\!\cdot\left[k(T)\nabla T\right]+S_H,\\
\frac{\partial \alpha}{\partial t}
+\nabla\!\cdot(\alpha\mathbf{u})&=0,\\
H&=h(T)+f_lL,\qquad
\boldsymbol{\tau}=\mu(T,\dot{\gamma})
\left(\nabla\mathbf{u}+\nabla\mathbf{u}^{\mathsf T}\right).
\end{aligned}
\label{eq:conservation-model}
\end{equation}
Here, $\rho$, $\mathbf{u}$, $p$ and $T$ denote density, velocity, pressure and temperature. The stress $\boldsymbol{\tau}$ permits variable viscosity, $H$ contains sensible enthalpy $h$, liquid fraction $f_l$ and latent heat $L$, and $\alpha$ tracks the free surface. Regular domains are discretised with D2Q9 or D3Q19 lattice populations, while unstructured geometries use a conservative finite-volume balance
\begin{equation}
\begin{aligned}
f_i(\mathbf{x}+\mathbf{c}_i\Delta t,t+\Delta t)-f_i(\mathbf{x},t)
&=-\omega\left(f_i-f_i^{\mathrm{eq}}\right)
+\Delta t\left(1-\frac{\omega}{2}\right)\mathcal{F}_i,\\
f_i^{\mathrm{eq}}
&=w_i\rho\left[1+\frac{\mathbf{c}_i\!\cdot\!\mathbf{u}}{c_s^2}
+\frac{(\mathbf{c}_i\!\cdot\!\mathbf{u})^2}{2c_s^4}
-\frac{\mathbf{u}\!\cdot\!\mathbf{u}}{2c_s^2}\right],\\
\frac{\mathrm{d}(V_c\mathbf{U}_c)}{\mathrm{d}t}
+\sum_{f\in\partial c}\left(\mathbf{F}^{\mathrm{adv}}_f
-\mathbf{F}^{\mathrm{diff}}_f\right)\!\cdot\!\mathbf{n}_f A_f
&=V_c\mathbf{S}_c .
\end{aligned}
\label{eq:fastfluent-discretisation}
\end{equation}
In Eq.~\ref{eq:fastfluent-discretisation}, $f_i$ is the population along lattice velocity $\mathbf{c}_i$, $\omega$ is the local relaxation rate and $\mathcal{F}_i$ is the discrete forcing term. The finite-volume state $\mathbf{U}_c$ is updated by advective and diffusive face fluxes over control volume $V_c$. Both formulations recover the physical fields used in the evidence object in Eq.~\ref{eq:evidence_object}. Supplementary Note~1 gives the macroscopic recovery, viscosity relation, bounded phase transport and conservation residuals used by FastAgent.

\subsection{Validation cases, controls and efficiency metrics}

Six public CFD benchmark campaigns covered channel flow, separated flow, blocked thermal transport, a lid-driven cavity, the Ghia cavity profiles and the Sch\"afer--Turek cylinder wake. Scalars used relative deviation, profiles used root-mean-square error and near-zero references used absolute error. The two literature reproductions follow Ghia and Sch\"afer--Turek \citep{ghia1982high,schaefer1996benchmark}. For case $i$ and policy $m\in\{\mathrm{FA},\mathrm{GA}\}$,
\begin{equation}
\begin{aligned}
\mathcal{E}_{i}^{(m)}
&=\max_j\frac{\left|q_{i,j}^{(m)}-q_{i,j}^{\mathrm{ref}}\right|}{\tau_{i,j}},\\
I_i&=100\left(1-\frac{\mathcal{E}_{i}^{(\mathrm{FA})}}
{\mathcal{E}_{i}^{(\mathrm{GA})}}\right)\%,
&S_i&=\frac{T_i^{(\mathrm{GA})}}{T_i^{(\mathrm{FA})}}.
\end{aligned}
\label{eq:agent-control-metrics}
\end{equation}
$\mathcal{E}_{i}^{(m)}$ is the normalised worst-QoI error, $I_i$ its reduction relative to the matched Generic Agent control and $S_i$ the decision-relevant speedup. Matched-evidence replay evaluated stagewise actions with the same campaign specification, candidates, progressively revealed CFD results, foundation-model environment and response format. In the timed workflow comparison, the Generic Agent used general scientific tool use with the fixed \fluent{} setup, whereas FastAgent combined its physics-aware policy with \fastfluent{} screening. Timings included screening and high-fidelity computation. Fixed \fluent{} schedules additionally supplied the solver-cost baseline in Fig.~\ref{fig:accuracy-efficiency}b. Supplementary Tables~S10 and S11 report the control definitions and stagewise decisions.

\subsection{FastAgent-guided startup screening and response measures}

The startup campaign resolved 0--4.1 s of steam admission. Three condensation treatments were compared against the imposed pressure rise and saturation-controlled thermal state \citep{minkowycz1966condensation,wagner2000iapws}. The selected treatment was applied to six orientation--distance cases and three additional inlet configurations. \fluent{} wall heat input was integrated through a lumped energy balance for a 5 mm ceramic shell. Spatial temperature departure and positive normal pressure supplied the thermal and impact components of the $P_{99.5}$ crack-driving response. The result was normalised by a 3.6 MPa green-shell reference, and a dense deterministic gain map tested ranking stability. Parameters are listed in Supplementary Tables~S4--S6.

\subsection{FastAgent-guided full-cycle screening and fidelity allocation}

The full-cycle \fluent{} model solved transient mass, momentum and energy transport over 0--420 s with gravity, temperature-dependent wax properties and enthalpy-porosity phase change \citep{piwonka2000modeling,sabau2003material}. For rapid candidate and resolution screening, \fastfluent{} solved transient conduction with an effective-heat-capacity representation of melting
\begin{equation}
\begin{aligned}
\rho c_{p,\mathrm{eff}}(T)\frac{\partial T}{\partial t}
&=\nabla\!\cdot\!\left[k(T)\nabla T\right],
&-k\nabla T\!\cdot\!\mathbf{n}&=h(T-T_{\mathrm{s}}),\\
f_l(T)&=
\begin{cases}
0, & T\leq T_{\mathrm{sol}},\\
(T-T_{\mathrm{sol}})/(T_{\mathrm{liq}}-T_{\mathrm{sol}}),
& T_{\mathrm{sol}}<T<T_{\mathrm{liq}},\\
1, & T\geq T_{\mathrm{liq}},
\end{cases}
&c_{p,\mathrm{eff}}&=c_p+L\,\frac{\mathrm{d}f_l}{\mathrm{d}T}.
\end{aligned}
\label{eq:dewaxing-thermal-screen}
\end{equation}
Here, $T_{\mathrm{s}}$ is the imposed steam-side temperature and $h$ is the convective heat-transfer coefficient. Harmonic face conductivity is used in the cell update. Liquid cells connected to the drainage boundary define the drainage accessibility $A_d$, which enters the screening pressure-risk proxy
\begin{equation}
A_d(t)=\frac{\left|\mathcal{C}_{\mathrm{out}}\{c:f_{l,c}\geq f_{\mathrm{crit}}\}\right|}{N_{\mathrm{wax}}},
\qquad
R_p(t)=\overline{f_l}(1-A_d)
\left(\frac{\overline{q}_w}{q_{w,0}}\right)^{0.45}
\left(1+0.25f_{\mathrm{mushy}}\right).
\label{eq:dewaxing-risk-proxy}
\end{equation}
This proxy combines melt accumulation, outlet-connected liquid, wall heat input and the fraction of cells within the melting interval. FastAgent uses its peak time for rapid candidate ranking, with \fluent{} supplying the corresponding high-fidelity evidence. The candidate objective is
\begin{equation}
J=0.60e_{\mathrm{melt}}+0.30e_{\mathrm{risk}}+0.10e_E
+0.05I_{\mathrm{no\,melt}}.
\label{eq:dewaxing-objective}
\end{equation}
The error terms describe full-melt time, risk-window time and energy balance, while $I_{\mathrm{no\,melt}}$ penalises incomplete melting. The 16-candidate search varied reduced-order physical parameters. A subsequent five-route comparison varied spatial and temporal resolution around the baseline route to determine QoI-specific numerical fidelity. These criteria allowed rapid risk-window localisation and separate \fluent{} confirmation of melt completion. Supplementary Table~S7 reports the criteria and selected routes.

\subsection{Structural and fracture analysis}

The measured pressure--volume--temperature relation converted \fluent{} temperature histories into retained wax expansion. Wax compressibility and shell compliance distributed this expansion over 16 shell regions according to
\begin{equation}
\left[\operatorname{diag}(\beta V)+\eta C_{\mathrm{shell}}\right]\mathbf{p}
\simeq \Delta\mathbf{V}_{\mathrm{retained}},
\qquad \mathbf{p}\geq 0,
\label{eq:regional-load-reconstruction}
\end{equation}
where $\beta=5.5\times10^{-4}$ MPa$^{-1}$ is the wax compressibility and $\eta=0.01$ defines the high-restraint reference state \citep{lee2016understanding}. The resulting regional loads drove Abaqus XFEM calculations of crack initiation, connectivity, cohesive damage and opening. Independent ProCAST contact pressures supplied a cross-solver spatial check. All structural outputs used a common physical time measured from the onset of heating. Supplementary Table~S8 reports the load construction and fracture response. Both global and local calculations remained quasi-static, and the 35/45 mm local-domain comparison produced essentially unchanged crack timing and opening. Energy-balance and domain-size results are reported in Supplementary Fig.~S12.

\subsection{FastAgent-guided outlet-layout screening and selection}

Seventeen binary outlet locations generated 131,072 layouts. \fastfluent{} screened all layouts, eleven underwent high-fidelity CFD evaluation with \fluent{} and three were examined structurally with Abaqus. Geometry, materials, supports and output schedules remained fixed, and only the pressure history changed. FastAgent retained normalised global drainage utility and local fracture safety as separate objectives. Fracture severity combined opening, cohesive damage, connected extent and onset timing with weights of 0.40, 0.25, 0.20 and 0.15, and its complement defined the safety score. A weight sweep in increments of 0.005 tested selection stability. Local exposure and opening ranks were compared using Spearman's rank correlation at 8.88 s. Supplementary Tables~S7--S9 provide the numerical and structural definitions.

\section{Conclusions}

FastAgent advances industrial CFD through the integration of \fastfluent{}'s rapid numerical screening, QoI-dependent fidelity control and discrepancy-updated design selection. The physical model, numerical resolution and simulation horizon adapt to the evidence needed for each decision. Across six public CFD benchmark campaigns, the framework reduces normalised worst-QoI error by 5.68--99.14\% and accelerates decision-relevant computation by factors of 1.53--15.36 relative to a matched Generic Agent control.

In steam dewaxing, the rapid route localises the full-cycle risk window using 81.39\% fewer cell-time updates than the fine-grid, small-time-step reference, while \fluent{} confirms melt completion. Complete \fastfluent{} screening of 131,072 layouts followed by selective CFD and structural evaluation identifies Sparse-6, which reduces simulated crack opening by 96.0\% and connected crack extent by 66.7\% relative to the Original design. This numerical--decision coupling connects large-scale exploration to mechanism-guided industrial design.

\section*{Data availability}
The data supporting this study are provided in the Article and Supplementary Information. Commercial confidentiality prevents public release of the case-specific industrial geometries, commercial-solver models and proprietary process data. Further information may be obtained from the corresponding authors upon reasonable request and subject to applicable confidentiality requirements.

\section*{Code availability}
The \fastfluent{} code and agent workflow are openly available at \nolinkurl{https://github.com/Yan-Kai-Chen/An-Agent-Framework-for-FastFluent-to-Fluent-Simulation-of-Complex-Solid-Liquid-Dynamics}.

\section*{Acknowledgments}
This work was supported by State Key Laboratory of Clean and Efficient Turbomachinery Power Equipment (Grant No. DEC8300CG202519639A1228342).

\section*{Author contributions}
Chen YanKai conceived the study, developed the workflow, performed the simulations and analyses, and wrote the manuscript. Yu QiuHeng contributed to simulation-model preparation, validation and interpretation. Yuan Man contributed to \fastfluent{} development. Zou Ting assisted with modelling and discussion. Wu Lin and He Jian provided case definitions and experimental observations. Xu QingYan, Jing Tao and Zhang QiongYuan supervised the study and revised the manuscript. All authors reviewed and approved the manuscript.

\section*{Competing interests}
The authors declare no competing interests.

\bibliographystyle{unsrtnat}
\bibliography{refs}

\clearpage
\thispagestyle{empty}
\mbox{}
\newpage
\includepdf[pages=-,pagecommand={\thispagestyle{empty}}]{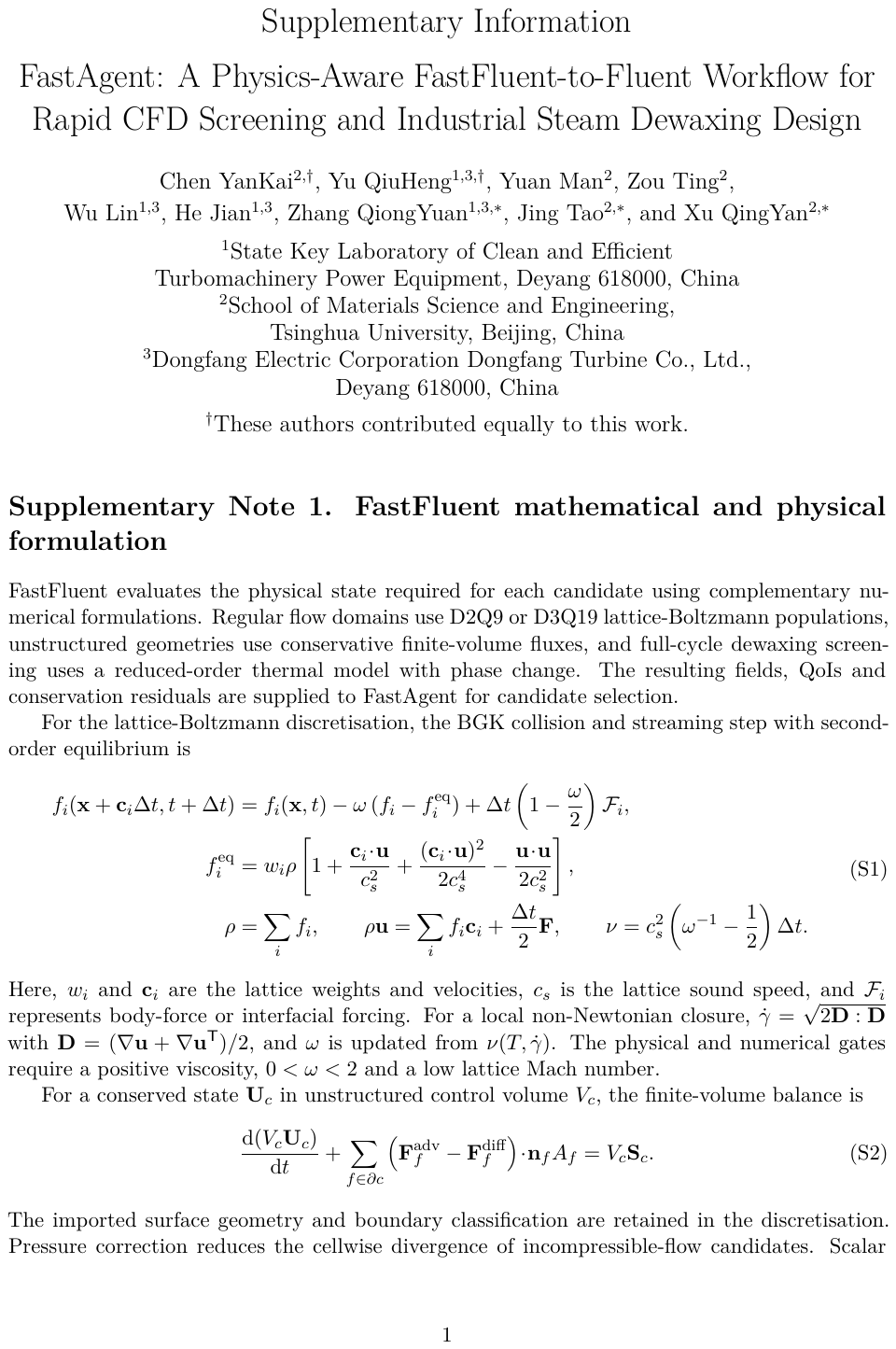}

\end{document}